\pdfoutput=1

\documentclass[11pt]{article}

\usepackage{acl}

\usepackage{times}
\usepackage{latexsym}
\usepackage{enumitem}

\usepackage{graphicx}
\usepackage{bbding}
\usepackage{pifont}
\usepackage{hyperref}
\usepackage{cleveref}
\usepackage{multirow}
\usepackage{lineno}
\usepackage{booktabs}
\usepackage{tabularx}

\def\eg{\emph{e.g.}} 
\def\ie{\emph{i.e.}}

\usepackage[T1]{fontenc}

\usepackage[utf8]{inputenc}

\usepackage{microtype}

\usepackage{inconsolata}

\title{Towards Objectively Benchmarking Social Intelligence \\for Language Agents at Action Level}

\author{Chenxu Wang $^{1,2}$, Bin Dai$^2$, Huaping Liu$^1$, \and Baoyuan Wang$^2$ \\
  $^1$Department of Computer Science and Technology, Tsinghua University ~ $^2$ Xiaobing.AI\\
  \texttt{wcx21@mails.tsinghua.edu.cn, daibin@xiaobing.ai,} \\ \texttt{hpliu@tsinghua.edu.cn, wangbaoyuan@xiaobing.ai} \\
  \\
  }
  
\begin{document}

\twocolumn[{
\renewcommand\twocolumn[1][]{#1}
\maketitle
\vspace{-15mm}
\begin{center}
    \captionsetup{type=figure} 
    \includegraphics[width=\textwidth]{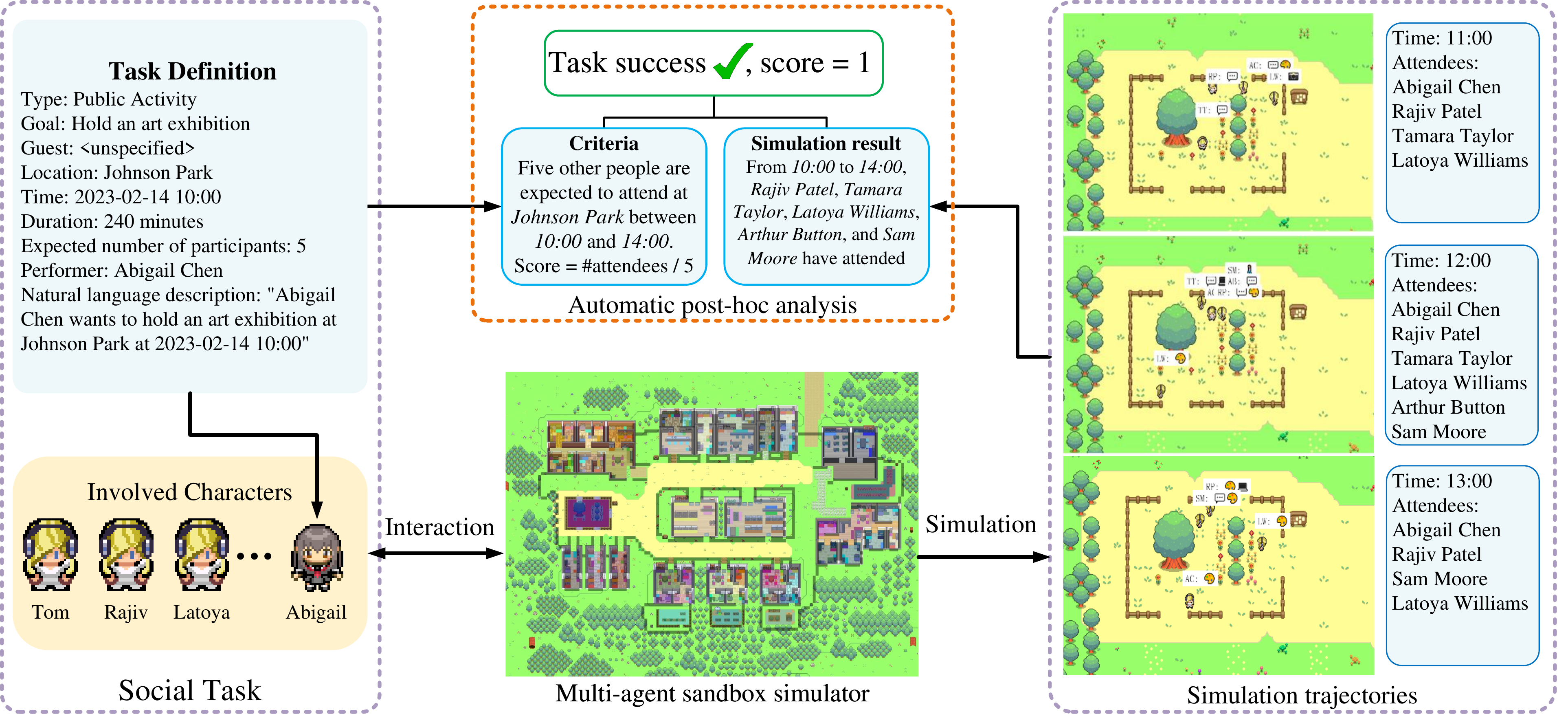}
    \caption{The general process of evaluating the language agent with STSS. At the beginning of each episode, a social task is assigned to a specific agent, \eg, \textit{Abigail wants to hold an art exhibition at Johnson Park}. The success of the task will be quantitatively measured with the trajectory in the simulation, \eg, 5 other agents attend the exhibition on time, signifying a full success of the task.}
    \label{fig: broad_idea}
    \vspace{5mm}
\end{center}
}]

\begin{abstract}
Prominent large language models have exhibited human-level performance in many domains, even enabling the derived agents to simulate human and social interactions. While practical works have substantiated the practicability of grounding language agents in sandbox simulation or embodied simulators, current social intelligence benchmarks either stay at the language level or use subjective metrics. In pursuit of a more realistic and objective evaluation, we introduce the Social Tasks in Sandbox Simulation (STSS) benchmark, which assesses language agents \textbf{objectively} at the \textbf{action level} by scrutinizing the goal achievements within the multi-agent simulation.
Additionally, we sample conversation scenarios to build a language-level benchmark to provide an economically prudent preliminary evaluation and align with prevailing benchmarks. To gauge the significance of agent architecture, we implement a target-driven planning (TDP) module as an adjunct to the existing agent. Our evaluative findings highlight that the STSS benchmark is challenging for state-of-the-art language agents. Furthermore, it effectively discriminates between distinct language agents, suggesting its usefulness as a benchmark for evaluating both language models and agent architectures.

\end{abstract}


\section{Introduction}
Large language models (LLMs) have become increasingly powerful owing to the escalation in model magnitude and technical advancements. Recent arts such as the GPT series \citep{brown2020language, ouyang2022training, openai2023gpt4} showcase human-level performance in diverse tasks within the natural language domain. Capitalizing on the instruction-following and emerging commonsense ability of LLMs, the derived language agents also demonstrate human-level ability in various scenarios~\citep{argyle2023out, aher2023using, Park2023GenerativeAgents, li2023metaagents, wang2023voyager}.

The endeavor to craft comprehensive language agents to simulate humans has prompted substantial attention toward the evaluation of social intelligence.
In recent times, numerous simulators have been developed for the social simulation of language agents~\citep{park2022social, Park2023GenerativeAgents, wang2023humanoid, li2023metaagents}. However, the challenge of quantitatively evaluating the emergent social intelligence in these simulations persists. While previous works have introduced numerous benchmarks and evaluation metrics~\citep{zhou2023sotopia, li2023metaagents}, we contend these evaluations face two fundamental issues. Firstly, these evaluations predominantly focus on the language level, assessing social intelligence through posed questions. However, it remains conceivable that an agent may claim they will perform a certain action without actually committing to it, rendering language-level evaluations inadequate for reflecting actual behaviors. Secondly, many evaluations hinge on subjective metrics, diminishing the reliability of the evaluation results. We argue that designing a social intelligence benchmark that can be evaluated \textbf{objectively} at the \textbf{action level}, rather than subjectively at the language level, is necessary.

In this paper, we introduce the Social Tasks in Sandbox Simulation (abbreviated as STSS), an action-level benchmark that assesses task-oriented social intelligence with objective metrics based on the Smallville environment \citep{Park2023GenerativeAgents}. We craft 30 templates of social tasks across 5 categories with corresponding mechanisms to automatically evaluate agents within the simulation. The overarching architecture of the STSS benchmark is illustrated in \Cref{fig: broad_idea}, where the agent is initiated with a social task and is measured through goal achievement in the simulation. 
Since the simulation may be economically expensive, we additionally construct a complementary language-level benchmark by capturing chat scenarios within the simulator, as a preliminary benchmark that aligns with existing language-level benchmarks such as SOTOPIA \citep{zhou2023sotopia}. We conduct a comparative analysis between our proposed STSS benchmark and prior works in \Cref{tab: BenchmarkComparison}.

Notably, our benchmark evaluates language agents instead of merely language models, considering that the architecture of agents, \ie, how to use the language models is also important to social intelligence. To verify our hypothesis, we introduce an additional Target-Driven Planning (abbreviated as TDP) module that aims at social tasks for the generative agents \citep{Park2023GenerativeAgents}.
While our framework is versatile for evaluating language models within a fixed baseline agent architecture, it concurrently serves as a testbed of language agent architectures.

We conduct evaluations with various popular LLMs on both levels of our benchmark. The results reveal that the social tasks remain challenging even for cutting-edge models like GPT-4 \cite{openai2023gpt4}, underscoring the potential space for improvement in language models. Moreover, a well-designed TDP module demonstrates a substantial enhancement in the performance of language agents at both levels, suggesting that our benchmark can also serve as a testbed of agent architectures employing fixed LLM.

Our contributions can be summarized as follows:
\begin{itemize}[leftmargin=10pt]
\vspace{-2mm}
    \item We introduce STSS, a two-level benchmark for evaluating the social intelligence of language agents, encompassing an action-level evaluation in sandbox simulations and a preliminary language-level evaluation in interactive conversations.
    \vspace{-2mm}
    \item We design a target-driven planning module for language agents to investigate the influence and importance of agent architecture in executing social tasks.
    \vspace{-2mm}
    \item We conduct extensive experiments with several state-of-the-art models to gauge the capabilities of existing language agents. The result also suggests the effectiveness of our benchmark in evaluating agent architecture.
\end{itemize}

\begin{table*}\small
    \centering
    \scalebox{0.9}{
    \begin{tabular}{cccccc} 
    \hline
     & Social Sim. & Target-oriented & Interactive & Sim. Level & Evaluation \\  
    \hline
    Social-IQ \citep{zadeh2019social} & \ding{55}  & N/A  & \ding{55} & Language & Objective  \\ 
    Social IQA \citep{sap2019socialiqa} & \ding{55}  & N/A  & \ding{55} & Language & Objective  \\ 
    ToMi \citep{le2019revisiting} & \ding{55}  & N/A  & \ding{55} & Language & Objective  \\ 
    HALIE \citep{lee2023evaluating} & \ding{55}  & \ding{55}  & \ding{51} & Language & Subjective  \\ 
    Cooperative Agent \citep{zhang2023building} & \ding{55}  & \ding{51} & \ding{51} & Embodied & Objective \\ 
    
    Generative Agents \citep{Park2023GenerativeAgents} & \ding{51}  & \ding{55}  & \ding{51} & Sandbox & Subjective  \\ 
    SOTOPIA \citep{zhou2023sotopia} & \ding{55}  & \ding{51}  & \ding{51} & Language & Subjective  \\ 
    MetaAgents \citep{li2023metaagents} & \ding{51}  & \ding{51}  & \ding{51} & Language & Objective  \\ 

    \hline
    STSS (ours) & \ding{51} & \ding{51} & \ding{51} & Sandbox & Objective \\
    \hline
    \end{tabular}
    }
    \caption{Comparison between our benchmark and related social Intelligence benchmarks. \textbf{Sim.} denotes simulation. We regard interaction among more than two agents in a shared space as social simulation.  We adopt the conceptual framework of subjective and objective evaluation from \citet{wang2023survey}. Alignment to human values is considered subjective, whereas choice questions, goal achievements, and rule-based are classified as objective.} 

    \label{tab: BenchmarkComparison}
\end{table*}

\section{Related Work}
    
\subsection{Social Intelligence Evaluation}

Social Intelligence (SI) is widely recognized as the ability to understand others and to act wisely in social situations \citep{walker1973social}. In the pursuit of benchmarking social intelligence, previous arts such as SOCIAL IQA \citep{sap2019socialiqa} and Social-IQ \citep{zadeh2019social} evaluate the model through question answering (QA) within given contexts. Recent works either proposed specialized benchmarks~\citep{le2019revisiting} for SI or evaluated SI on LLMs~\citep{sap2022neural, shapira2023clever, shapira2023well}, both revealing imperfections of current LLMs.

Moving beyond static and non-interactive QA, recent research advocates evaluating LLMs in interactive benchmarks \citep{lee2023evaluating, zhou2023sotopia}. 
Originating from social requirements, target-oriented dialogue has been investigated both in terms of model and benchmark \citep{wang2019persuasion, hosseini2020simple, meta2022human, tiwari2023towards}. Moreover, numerous studies also address the quality of conversations, such as the personalization of the chatbot \citep{zhang2018personalizing, liu2020you, ait2023power}. Taking a step further, SOTOPIA \citep{zhou2023sotopia} introduces explicit social goals/persona on the agents.

However, existing benchmarks either operate solely at the language level or rely on subjective evaluation metrics. In contrast, our STSS benchmark introduces an objective metric that assesses both language and grounded actions, thereby enhancing the reliability of the evaluation.

\subsection{Language agents}

There has been a growing interest in constructing language agents by integrating them with external environments and anchoring the language to actions~\citep{andreas2022language, wang2023survey}. Recent studies demonstrate that language agents can undertake complex tasks, including game playing \citep{wang2023voyager}, acting as research assistants \citep{ziems2023can, bran2023augmenting}, robotic planning \citep{huang2022language, brohan2023can}, and even simulating humans \citep{aher2023using}. 
A mass of benchmarks and environments have been built for evaluating the language agents. ALFWorld \citep{shridhar2020alfworld} established the bridge between language and embodied robotics tasks. AgentBench \citep{liu2023agentbench} provides a comprehensive benchmark including 8 tasks. Motivated by the human-like social behaviors of the language agents, there also have been social simulators allowing multi-agent interaction in a shared environment \citep{park2022social, Park2023GenerativeAgents, gao2023s3, li2023metaagents}.

It has been substantiated that the utilization of LLMs can significantly influence the performance of language agents. Chain-of-Thought(CoT)~\citep{wei2022chain}, along with various other works~\citep{yao2023tree,besta2023graph,chia2023contrastive,nori2023can} demonstrates the significance of prompt designation to the performance of LLMs. Advanced agent architecture such as ReAct~\citep{yao2022react} and Relfexion~\citep{shinn2023reflexion} further improve the performance in both pre-hoc and post-hoc ways ~\citep{huang2023large}. BOLAA~\citep{liu2023bolaa} systematically investigates agent architectures and concludes that a well-designed architecture with dedicated modules for specific tasks can significantly enhance agent performance. For human simulation, Generative Agents~\cite{Park2023GenerativeAgents} designs a complete pipeline and preliminarily verifies the idea of social agents.

While simulations have provided a decent testbed of social behavior, few works attempt to objectively evaluate the social intelligence of language agents quantitatively. Our STSS benchmark fills the gap between social intelligence evaluation and human behavior simulation, with the potential to generalize to more realistic scenarios such as embodied tasks and even real-world applications.

\begin{table*}
    \small
    \centering

    \begin{tabular}{cccccc} 
    \hline
    Task Type & Time & Location & Target & Metrics & \#Templates \\  
    \hline
    Public activity & Specified & Specified & \ding{55} & \#Participants & 10 \\ 
    
    Appointment & Same & Same & \ding{51} & Goal achievement & 5 \\ 
    Inviting companions & Same & Specified & \ding{55}& \#Participants & 5 \\ 
    Online activity & Specified & Unspecified & \ding{55} & \#Participants & 5 \\ 
    Asking for help & Unspecified & Specified & \ding{55} & Goal achievement & 5 \\ 
    \hline
    \end{tabular}
    
    \caption{Detailed features of social task types in our benchmark. \textbf{Specific} denotes the time or location is predetermined, \textbf{Same} requires the agent to negotiate with others to fix the time or location, whereas \textbf{Unspecified} does not impose explicit restrictions on the time or location. \textbf{Target} indicates whether other characters are specified in the task, \eg, in \textbf{Appointment} tasks, the agent may need to talk with specific characters.} 
    
    \label{tab: BenchmarkDetails}
\end{table*}


\section{Social Tasks in Sandbox Simulation}

To objectively assess the social intelligence of language agents at the action level, we suggest building benchmarks within interactive simulators. The simulator can concretize the textual output into tangible actions, facilitating a more objective and quantitative evaluation through analysis of the state and trajectory of the simulation environment.

We extend the interactive sandbox environment and the generative agent architecture introduced by  \citet{Park2023GenerativeAgents} to accommodate the social intelligence evaluation. Specifically, we devise a set of social task templates and corresponding metrics, which can be quantified by analyzing the actions of the agents. Take the example of \textit{holding a Valentine's Day party}, the metric could be the number of participants at the correct time and location.
These templates are then instantiated to specific tasks, each associated with corresponding initial states of the environment, being available for assessing language agents within the simulation.

\subsection{Task Designation}
\label{sec: task}
We aim to evaluate several fundamental capabilities in social interaction, including \textbf{planning}, \textbf{expression}, and \textbf{negotiation}. Confronting with a social task, the agent naturally needs to formulate a comprehensive plan and execute necessary actions such as guest invitations and delivering key messages during conversations, requiring the ability of \textbf{planning} across various domains. 
When conversing with others, the capabilities of \textbf{expression} and \textbf{negotiation} exhibit their significance. The agent not only needs to \textbf{express} their thoughts and convey information sufficiently and clearly but also needs to \textbf{negotiate} with others when encountering matters to be determined. Any omission or miscommunication could hamper the success of the task, even the other agent might still make verbal consent. 


Driven by the need for objective action-level evaluation, the design of task templates focuses on the triad of \textbf{who}, \textbf{when}, and \textbf{where}. Broadly, we design task templates that are categorized into five distinct types facing different scenarios with diverse constraints:

\begin{enumerate}[leftmargin=10pt]
    \vspace{-2mm}
    \item \textbf{Public activity} with a fixed time and a fixed location, such as a Valentine's Day party or a public lecture. The agent is required to invite as many other agents as possible to the activity, explicitly expressing the designated time and location in each invitation. The objective metric is the number of participants who attend at the correct location during the specified time.
    \vspace{-2mm}
    \item \textbf{Appointment} with unconstrained time and location, \eg, a student is trying to schedule a discussion with the professor. The agent needs to locate the target person and arrange the appointment to accomplish the designated purpose. The challenge lies in negotiating with the target person to figure out a convenient time and location for both agents through conversation. The criterion is whether they execute the required action at the same place and time.
    \vspace{-2mm}
    \item \textbf{Inviting companions} for an activity with unconstrained time and a fixed location, such as gathering friends for a shopping outing. The agent is tasked with inviting others to the activity and negotiating the timing. The metric assesses how many participants joined the activity at the correct location, with the initiator being a mandatory attendee.
    \vspace{-2mm}
    \item \textbf{Online activity} with a fixed time and no location requirement, encompassing scenarios like an online workshop or playing online games together. In contrast to public activities, online activities do not necessitate participants to appear in a specific physical location.
    \vspace{-2mm}
    \item \textbf{Asking for help}, incorporating a series of activities at fixed locations that have no time constraint. In this category of tasks, the agent is assumed to be occupied or unable to do something, seeking assistance from others. For example, Wolfgang, facing an upcoming exam, needs a friend to borrow books from the library and deliver the books to his dormitory. The criterion evaluates whether other people execute the specific behavior correctly, \eg, first \textit{borrowing books} at the \textit{library} and then \textit{delivering the books} at the \textit{dormitory of the student}. 
\end{enumerate}

We incorporate a total of 30 social activity templates in the benchmark across the five categories, where more details of the categories are elaborated in \Cref{tab: BenchmarkDetails}. The task templates designate the task type, activity name and description, and other task-related information. They can then be instantiated to specific tasks by determining the variables such as the background characters. 
To simplify, we evaluate only one social task in each complete simulation, where the task is assigned to a single agent at the beginning and the other background characters remain agnostic to the task.
To manage the computational cost, we only sample 10-12 background characters in each simulation.  

\paragraph{Metrics} The performance is assessed through the trajectories in the simulation. Each type of task is associated with a set of rules to analyze the agent trajectories including their actions and positions, and generate a quantitative score to measure the success of the social task. 
For collective activities including \textit{public activity}, \textit{online activity}, and \textit{inviting companions}, scores are determined based on the ratio between the actual and expected number of participants. For \textit{appointment}, the agent receives a full score if the goal was perfectly achieved and a half score if the location is mismatched. \textit{For asking for help}, the score is the ratio of correctly performed sub-tasks. More details can be found in the appendix. 
For all the tasks, a keyword filter is applied to ensure that agents execute relevant actions and filter out irrelevant agents who accidentally pass by.


We clarify that our benchmark presently concentrates on objective and event-oriented evaluation, \ie, we only take the success of the task into account, excluding the measurement of other aspects such as caring about subjective feelings, personalization, or higher-level motivations. We leave the holistic evaluation of both subjective feelings and high-level social goals as future work.




\begin{figure*}[t]
\centering
\includegraphics[width=\textwidth]{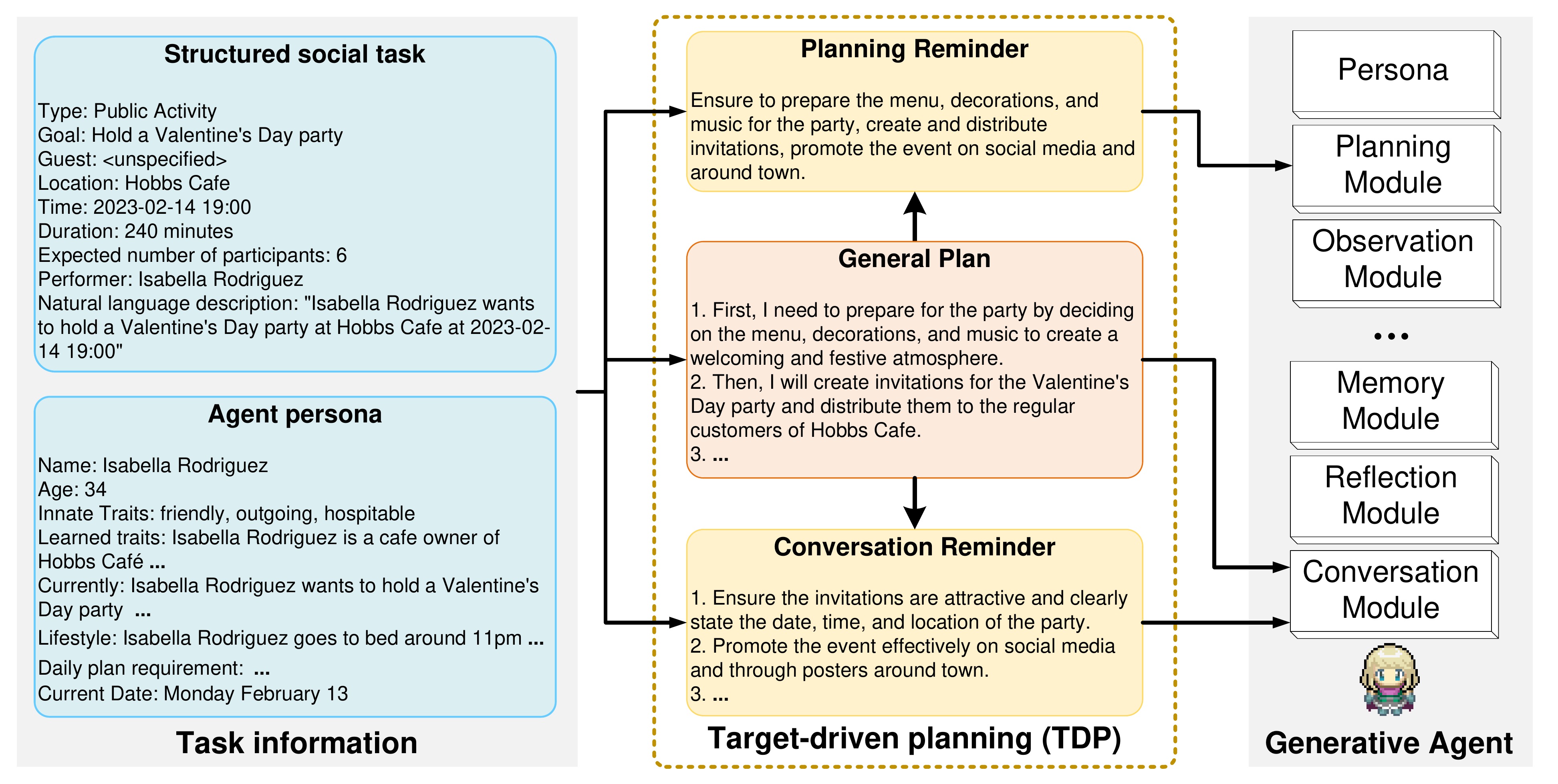}
\caption{An illustration of the TDP module attached to the generative agent. The generated plans and thoughts will be injected into the corresponding module of the agent.}
\label{fig: task_driven_planning}
\end{figure*}

\subsection{Language-Level Benchmark}
While comprehensive simulation-based evaluation is effective and objective, it may be expensive in terms of both time and economic sense. Besides, implementing a language agent to simulate human-like behavior in environments such as generative agents \citep{Park2023GenerativeAgents} may impose a relatively high minimum requirement on the capabilities of the language model. This hinders the universal availability of the benchmark. To mitigate this constraint, we suggest building an accompanying benchmark that focuses on situational dialogue, which is a critical capability for social intelligence. 

A conversation task is defined as a tuple (\textit{task}, \textit{performer}, \textit{target character}) and is curated by freezing the state from simulations. We collect 325 conversation scenarios from numerous simulations of 30 tasks instantiated from the templates. To ensure diversity, each task includes at most two conversation instances for the same pair of characters (although scenarios may already have different background information). For appointment tasks, only conversations between the task performer and the target person are incorporated into the benchmark.
In contrast to previous conversation benchmarks \citep{zhou2023sotopia, lee2023evaluating}, our conversation tasks are sliced samples from complete simulations and the characters are associated with self-consistent memories in the simulation worlds. These memories provide more diversity and naturalness than purely synthetic tasks. 

\paragraph{Metrics} To better align the conversation tasks with the simulations, we adopt an event-oriented assessment for the generated conversations. For each type of task, a set of goal achievement conditions is designed to measure the conversation, \eg, correctly conveying or reaching a consensus on the time or location of the event. The measurement of such conditions resembles reading comprehension problems and can be solved by language models. Compared to grading the conversation in continuous space, the goal conditions present binary classification problems, which can be easily aligned to human evaluation. 

We introduce two metrics in the situational conversation benchmark: 
\begin{itemize}[leftmargin=10pt]
    \vspace{-2mm}
    \item Success rate (SR). A task is considered successful if and only if the agent achieves all the goal conditions (for example, conveying all the key information such as time and location). The score is either 1 or 0 for each task.
    \vspace{-2mm}
    \item Goal condition SR (GCSR). We borrow the idea of goal-condition success from embodied vision-and-language tasks \citep{shridhar2020alfred}, where the score is the ratio of achieved conditions. For example, in the public activity tasks, if the agent (1) successfully makes the invitation; (2) appropriately informs the location; and (3) forgets to inform the time, then the goal-condition score will be $2/3 \approx 0.667$.
\end{itemize}

In consideration of the practical application, the produced conversation is expected to be not only informative but also fluent, clear, and concise. To encourage more effective conversations, we additionally adopt summary-level evaluation, where the conversion is first summarised by the other person (act by an improved version of the generative agent), and then the same metrics are applied to evaluate whether the summary includes the key information. The summary-level evaluation additionally inspects the quality of utterance. For example, even if containing key information, chaotic or blurry expressions may hinder the listeners from correctly summarizing the information, let alone performing the actions.  

We also note that the number of conversations can differ sharply across tasks, \eg, there will be only one valid conversation in appointment tasks, but a task where a social butterfly is holding a party may include up to 10 conversations. To enrich the dimension of evaluation, we employ micro average and macro average for the scores. Let $\mathcal{T}$ be the set of tasks, $\mathcal{C}_{t}$ denote the conversation set of task $t$, and $s_{c}$ be the score of a conversation $c$, the micro average score is defined as:
\begin{equation}\vspace{-0.2cm}
    \bar{s}_{micro} = \frac{\sum_{t \in T} \sum_{c \in \mathcal{C}_{t}} s_c}{\sum_{t \in T} \left |\mathcal{C}_{t} \right |}.
\end{equation}
And the macro score averages across tasks, being defined as:
\begin{equation}\vspace{-0.2cm}
    \bar{s}_{macro} = \frac{1}{\left |\mathcal{T} \right |} \sum_{t \in T} \frac{\sum_{c \in \mathcal{C}_{t}} s_c}{\left |\mathcal{C}_{t} \right |} .
\end{equation}

\begin{table*}\small
    \centering
    \begin{tabular}{cccccccccc} 
    \hline
    \multirow{3}{*}{LLM} & \multirow{3}{*}{Agent} & \multicolumn{4}{c}{Conversation} & \multicolumn{4}{c}{Summary} \\ 
    \cline{3-10}
    & & \multicolumn{2}{c}{SR} & \multicolumn{2}{c}{GCSR} & \multicolumn{2}{c}{SR} & \multicolumn{2}{c}{GCSR} \\
    \cline{3-10}
    & & Micro & Macro & Micro & Macro & Micro & Macro & Micro & Macro \\
    \hline
    Llama2 & GA & 0.212 & 0.190 & 0.382  & 0.366 & 0.167 & 0.139 & 0.324 & 0.272 \\ 

    Baichuan2 & GA & 0.437 & 0.407 & 0.615 & 0.636 & 0.231 & 0.226 & 0.460 & 0.438 \\ 
    
    GPT-35 & GA & 0.372 & 0.335 & 0.486 & 0.455 & 0.243 & 0.192 & 0.381 & 0.323 \\ 

    GPT-4 & GA & 0.615 & 0.557 & 0.748  & 0.715 & 0.486 & 0.402 & 0.656 & 0.569 \\ 
    \hline
    
    Llama2 & GA + TDP & 0.502 & 0.433 & 0.656 & 0.601 & 0.305 & 0.276 & 0.539 & 0.476 \\ 

    Baichuan2 & GA + TDP & 0.662 & 0.700 & 0.820 & 0.841 & 0.403 & 0.313 & 0.660 & 0.566 \\ 

    GPT-35 & GA + TDP & \underline{0.751} & \textbf{0.762} & \underline{0.862} & \underline{0.864} & \underline{0.570} & \underline{0.511} & \underline{0.734} & \underline{0.692} \\ 

    GPT-4 & GA + TDP & \textbf{0.812} & \underline{0.748} & \textbf{0.902} & \textbf{0.880} & \textbf{0.723} & \textbf{0.608} & \textbf{0.874} & \textbf{0.787} \\ 
    \hline
    \end{tabular}
    \caption{Overall performance across tasks of Language agents in the situational conversation.} 

    \label{tab: Results_L2}
\end{table*}

\begin{table*}\small
    \centering
    \begin{tabular}{ccccccccc} 
    \hline
    LLM & Agent & Pub. Act. & Appo. & Inv. Com. & Online Act. & Help & Overall & Conv. Ratio \\  
    \hline
    GPT-35 & GA & 0.338 & \underline{0.200} & 0.100 & 0.571 & 0.400 & 0.324 & \textbf{0.891} \\ 

    GPT-4 & GA & 0.590 & 0.000 & 0.000 & \underline{0.614} & \textbf{0.600} & 0.399 & 0.781 \\ 
    \hline
    GPT-35 & GA + TDP & \underline{0.608} & 0.100 & \textbf{0.200} & 0.471 & \textbf{0.600} & \underline{0.431} & \underline{0.866} \\ 

    GPT-4 & GA + TDP & \textbf{0.759} & \textbf{0.400} & \textbf{0.200} & \textbf{0.686} & 0.500 & \textbf{0.550} & 0.863 \\ 
    \hline
    \end{tabular}
    
    \caption{Average scores achieved within each task type and the overall average in the simulation. The abbreviations correspond to the 5 task types as introduced in \Cref{tab: BenchmarkDetails} and \textit{Conv. Ratio} stands for the ratio of initiated conversations to the number demanded in each task and may exceed 1 if the amount of conversion exceeds the requirement. Due to the characteristics of \textit{ask for help} tasks, we exclude them from the statistic of conversation amounts.} 

    \label{tab: Results_L1}
\end{table*}

\section{Task Driven Planning}
Our benchmark focuses on evaluating language agents rather than merely language models. Consistent with prior research findings \citep{liu2023bolaa}, we assert that designing specialist agent architecture is important to the application of language models. To enhance language models in social tasks, we propose an additional module namely target-driven planning (TDP) for the generative agent. We implement a simple baseline, as presented in \Cref{fig: task_driven_planning}, where the language model is first asked to provide a general plan, and then offer specific thoughts for daily planning and conversation. These thoughts are then injected into the corresponding module of the generative agent. The TDP module is also applicable to the accompanying conversation benchmark, although only the conversation module is active.

Our implementation does not incorporate advanced prompt techniques such as CoT \citep{wei2022chain} and
ReAct \citep{yao2022react}. We clarify that our implementation primarily serves as a baseline to inspect the significance of language agent architectures in performing social tasks. There remains ample room for improvement in the designation of agent architectures.

\section{Experiments}
In this section, we present the experiments conducted on our benchmarks to assess the social intelligence of language agents through performing social tasks. Our experimentation centers on addressing two key questions: (1) To what extent can existing language models and agents execute social tasks within our STSS benchmark? (2) What is the significance of the agent architecture in the context of social intelligence?

\subsection{Experiment Setup}
We incorporate 4 language models in the evaluation: GPT-35 \footnote{gpt-35-turbo-0613} \citep{ouyang2022training}, GPT-4 \footnote{gpt-4-0613} \citep{openai2023gpt4}, Llama-2-13b-chat \citep{touvron2023llama}, and Baichuan-2-13b-chat \cite{yang2023baichuan}.

For the agents, we utilize prompt templates and parameters of generative agents~\citep{Park2023GenerativeAgents} with minor adjustments for robustness in both levels of the benchmark. 
In the simulation, GPT-35 is employed as the language model for background characters, as this practice is known to provide human-level intelligence \citep{Park2023GenerativeAgents}. The evaluated model is used to generate target-driven plans, conversations, and high-level planning for the performer agent, while the low-level simulation is delegated to GPT-35.
For language-level evaluation, the task performer is operated by the evaluated language agent, while the others are handled by vanilla generative agents with GPT-35. We also use GPT-35 as the evaluator of conversation and summary.

Due to the minimum requirement of the instruction-following ability in the simulation, we exclusively GPT-35 and GPT-4 in the complete simulation. Since Llama2 and Baichuan2 exhibit instability when following the complex instructions from the generative agent, we simplify the prompt template of conversation generation in their evaluation. Please refer to the appendix for more details about the experimental setup to reproduce the empirical results.

In addition, to study the significance of the agent architecture, we conduct experiments on all models using the vanilla generative agent (GA) and the one with target-driven planning (GA + TDP). For GA, we adhere to the practice of \citet{Park2023GenerativeAgents}, injecting the social goal description into the \textit{currently} field of the agent persona.

\subsection{Overall performance}
We present the language-level evaluation results in \Cref{tab: Results_L2} and the action-level evaluation in \Cref{tab: Results_L1}.

In general, at the language level, GPT-4 performs the best, followed by GPT-35, Baichuan2, and Llama2. 
When evaluating the conversation, GPT-4 with TDP achieves a success rate of 81.2\% and a goal-condition success rate of 90.2\%, demonstrating its robust ability in target-oriented conversations. Surprisingly, Baichuan2 outperforms GPT-35 in the vanilla setting, although GPT-35 performs better with the TDP. We hypothesize that Baichuan2 possesses a better innate inferential capability for planning, whereas GPT-35 excels in generating higher-quality English utterances. 
Even though GPT-35 surprisingly performs comparably to GPT-4 with TDP in the conversation-level evaluation, it falls short when evaluating the summary, implying GPT-4 still produces higher-quality content. 
The consistent discrepancy between conversation and summary scores indicates a ubiquitous need for improvement in conversation quality.

While GPT-4 showcases commendable performance in the sampled situational dialogues, executing social tasks in a sandbox simulation remains challenging. Even the most proficient agent (GPT-4 with TDP) only achieves an average score of 0.550. When comparing task types, we find that making \textit{appointments} and \textit{inviting companions} emerge as the most challenging tasks. In these tasks, agents must inform others and negotiate regarding time or locations, necessitating the seamless integration of all abilities mentioned in Section~\ref{sec: task}.

\subsection{Comparing Language-Level and Action-Level Evaluations}

We observe a general consistency between the action-level and the language-level evaluations, as reflected in the positive correlation between the scores achieved by each agent. However, all agents experience a decrease in performance from language-level to action-level.
This supports our hypothesis that the sandbox simulation is more complex and challenging. By inspecting the number of initiated conversations, we identify threefold challenges: (1) The agent may not always meet enough other agents to chat with in the simulation, which depends on its itinerary and fundamentally its ability to plan; (2) The agent may struggle to effectively communicate information or determine an appropriate time and location during a conversation; (3) Even when information is accurately conveyed and understood, other agents may not always correctly ground it into actions.

Sandboxes can simulate the nature of the real world, exposing new drawbacks of language agents that can not be reflected by language-level evaluation alone. For instance, GPT-4 with TDP unexpectedly failed in an \textit{inviting companions} task where \textit{Wolfgang needs to invite friends to exercise with at Johnson Park}. In the simulation, Wolfgang achieved a verbal agreement with several agents to meet at Johnson Park tomorrow at 6:00 am. However, despite the appointment, most agents failed to wake up until 7:00 am the next day, resulting in the appointment being missed.



\begin{table}[t!]\small
    \centering
    \begin{tabularx}{1.0\linewidth}{ | >{\centering\arraybackslash}X| }
    \hline
    \textbf{Without TDP} \\
    \hline 
    Hi Tom, how's your day going? I'm planning a Valentine's Day party at the cafe tomorrow. Would you like to come? \\
    \hline\hline 
    \textbf{With TDP} \\
    \hline 
    Hi Tom, I hope you're doing well. I'm planning a Valentine's Day party \textbf{at Hobbs Cafe tomorrow at 7pm}. I would love it if you could come and \textbf{bring some friends along. The more, the merrier!} \\
    \hline 
    \end{tabularx}
    \caption{Isabella's invitations with/without the TDP module. When the TDP module is adopted, the invitation includes a clear date, time, and location of the party, facilitating the success of the social task.}
    \label{tab:invitation}
\end{table}

\subsection{Effectiveness of TDP}

For all language models, the GA + TDP architecture consistently outperforms the vanilla GA. The TDP module yields notable improvements in both levels of evaluation, and can even compensate for the shortage of model capabilities. For instance, GPT-35 and Baichuan2 agents with the TDP module can outperform the vanilla GA using GPT-4. Despite GPT-35 being empirically weaker than GPT-4, their conversation scores become almost comparable after equipping the TDP module. This suggests that when a language model exhibits capabilities above a certain threshold, the agent architecture may be the more noteworthy part for further improvement. 

The case of \textit{Isabella’s Valentine’s Day party at Hobbs Cafe}, previously presented in \Cref{fig: task_driven_planning}, provides insight into the performance boost facilitated by TDP. 
A comparison between invitations from Isabella with/without the TDP module is demonstrated in Table~\ref{tab:invitation}. The conversation reminder generated by the TDP prompts the agent to  \textit{clearly state the date, time, and location of the party}, which is vital for the success of the task. The invitation generated with the TDP also encourages the invitee to bring more friends, further facilitating the success of the social task. Please refer to the appendix for more details.

\section{Conclusion}
In this paper, we introduce the Social Tasks in Sandbox Simulation (STSS) benchmark, which is designed to assess language agents by engaging them in social tasks within a sandbox simulation. The accompanying language-level benchmark also serves as a preliminary evaluation for weaker models. Additionally, we propose a target-driven planning module for generative agents to investigate the significance of designing a specialized agent architecture in social intelligence. A comprehensive evaluation involving four prominent language models validates the efficacy of our benchmark for assessing both language models and agent architectures.


\clearpage
\section*{Limitations}

Though simulators offer numerous advantages, economic considerations pose a potential limitation. Based on the 2023 pricing of the OpenAI API, evaluating a single language agent on the 325 scenarios at the language level costs approximately \$7.5, whereas a comprehensive sandbox evaluation of 30 tasks costs around \$300. Due to the cost constraints, we do not conduct repetitive simulations, which could cause random fluctuations in the scores.

We also acknowledge that a gap exists between the sandbox simulation and the real world. Before migrating agents to realistic applications, it might be necessary to study the alignment between simulated performance and human evaluation.

\bibliography{custom}

\clearpage
\appendix
\label{sec:appendix}

\section{Experimental Details}
We follow the simulation framework introduced by Generative Agents~\cite{Park2023GenerativeAgents}, where the personas are generally initialized with a series of fields including: \textit{name}, \textit{age}, \textit{innate}, \textit{learned}, \textit{currently}, \textit{lifestyle}, and \textit{living area}. To prompt agents to perform the social tasks, we inject the task information into the \textit{currently} field of the agent, such as: \textit{Isabella Rodriguez wants to hold a Valentine's Day party at Hobbs Cafe at 2023-02-14 19:00. The target is to have 5 people attend the party at Hobbs Cafe at 2023-02-14 19:00, the more people the better}. In this way, as a part of the persona, the task information will appear in the prompts at various stages such as planning or conversation.

In the simulation, agents will make general plans, decompose their plans into language-level actions, determine the location of each action, and move to the designated locations when performing the actions. 
When drawing up the plan, agents also take their memories into account, including their observation in the simulation sandbox and summary of past conversations. As the most direct way of information interchange, conversation plays an important role in social simulation and is vital to the success of social tasks.

\begin{table*}[t]\small
    \centering
    \begin{tabular}{cp{12.5cm}} 
    \hline
    Task Description & Goal: Hold a Valentine's Day party at Hobbs Cafe at 2023-02-14 19:00 on 2023-02-14 19:00. Criteria: The target is to have 6 people attend the party at Hobbs Cafe at 2023-02-14 19:00, the more people the better. \\  
    \hline
    General Plan & 1. First, I need to prepare for the party by deciding on the menu, decorations, and music to create a welcoming and festive atmosphere. 
    2. Then, I will create invitations for the Valentine's Day party at Hobbs Cafe and distribute them to regular customers and potential new ones. 
    3. I will also use social media platforms to promote the event and attract more people. 
    4. On the day of the party, I will ensure everything is set up properly and ready for the guests. 
    5. Finally, I will host the party, making sure everyone feels welcome and has a good time. \\  
    \hline
     Plan Reminder & Ensure to prepare the menu, decorations, and music for the party, create and distribute invitations, and promote the event on social media platforms. \\
    \hline
     Conversation Reminder & Ensure the invitations are attractive and clearly state the date, time, and location of the party.
Promote the event effectively on social media, highlighting the unique features of the party.
Prepare a menu that caters to a variety of tastes and dietary preferences.
Decorate the cafe in a festive and welcoming manner.
Be a gracious host, ensuring all guests feel welcome and enjoy themselves. \\
\hline 
    \end{tabular}
    \caption{Full results of the target-driven planning in the case of Isabella's party.} 
    
    \label{tab: TDPCase}
\end{table*}

\subsection{Task Scoring}
For collective tasks including \textit{public activity}, \textit{inviting companions}, and \textit{online activities}, the score $s$ is granted regarding the proportion of participants and has an upper limit of $1$, formulated as:
$$ s = min(\frac{\#actual~participants}{\#expected~participants}, 1),$$
where the agent is expected to invite half of the background characters for \textit{public activity} and \textit{online activity}, and one-third of the background characters for \textit{inviting companions}; all numbers are rounded up.

In \textit{appointment}, the criterion is whether the agent and the invitee fulfill the appointment, \ie, performing the designated action at the same time and the same location. Only location mismatch will deduct half the score and other cases are regarded as fail, resulting in a score of 0. 

In \textit{asking for help}, the agent needs to ask others to perform a designated sequence of actions. The score is intended to be the proportion of actions that are performed in the appropriate order, nevertheless, since the \textit{asking for help} tasks in our datasets only have two subtasks, we simplify the criterion as follows: (1) Successfully perform all subtasks in the correct order will lead to a full score; (2) If not completely success, successfully performing at least one subtasks results in half score; (3) Otherwise, the score is 0.

\subsection{Conversation}
\begin{figure}[t]
\centering
\includegraphics[width=0.5\textwidth]{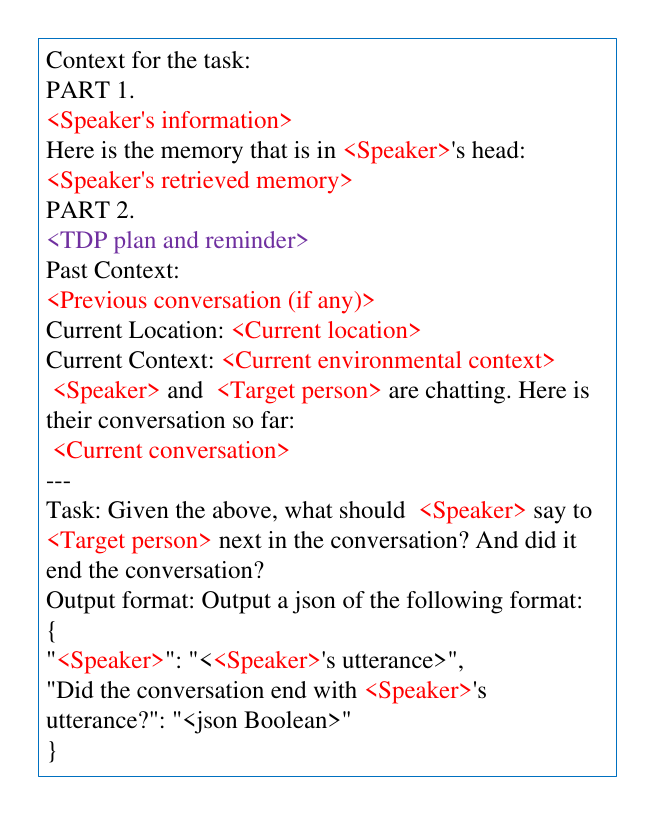}
\caption{Prompt template for the conversation, where angle brackets denote the placeholders that will be replaced by corresponding content when instantiating.}
\label{fig: ConversationPrompt}
\end{figure}

The template of conversation prompts is illustrated in \Cref{fig: ConversationPrompt}, generally including the character information, conversation context, and instructions for the LLM. When performing a social task, the relevant information will appear in the \textit{Speaker's information}. 

For the conversation, we set the temperature, frequency penalty, and presence penalty to 0. Each conversation lasts at most 8 rounds.
LLama2 and Baichuan2 are not prompted to output in the JSON format and are only asked to output the utterance. We parse their output as plain text and end the conversation when the round limit is reached.

\section{Target Driven Planning}
We keep the prompt style in the TDP module consistent with the generative agents. In all three steps, the character information and task information are organized in the same way, whereas the following instructions are slightly different, specified as the following:
\begin{enumerate}
    \item Generating the general plan: \textit{Here comes the first step of planning, you need to decompose the task into few subgoals or keypoints to fulfill the task requirements.
Output your thought in 3$\sim$5 sentences, one per line, without any prefix or postfix.}
    \item Generating the plan reminder: \textit{Here comes the next step of planning. When you are making the schedule of the day (<date>), what should you keep in mind?
Make a reminder which has 1$\sim$2 sentences, in one line, without any prefix or postfix.}
    \item Generating the conversation reminder: \textit{Here comes the next step of planning. When you are talking to others, what should you keep in mind to accomplish the goal?
Make a checklist with less than 5 items, one per line, without any prefix or postfix.}
\end{enumerate}


We present the results of each stage in the TDP module of a representative case, Isabella's invitation, in \Cref{tab: TDPCase}. For all generations in the TDP, the temperature is set to 0 for stable reproduction. 
As illustrated in \Cref{fig: task_driven_planning}, the \textit{plan reminder} will be provided when the agent plans for the new day, while the \textit{general plan} and the \textit{conversation reminder} will appear in the prompts for conversation, and \textit{task description} is applicable for both.

\end{document}